\documentclass[sigconf]{acmart}

\copyrightyear{2026}
\acmYear{2026}
\acmConference[MM '26] {Proceedings of the 34th ACM International Conference on Multimedia}{November 10--14, 2026}{Rio de Janeiro, Brazil.}
\acmBooktitle{Proceedings of the 34th ACM International Conference on Multimedia (MM '26), November 10--14, 2026, Rio de Janeiro, Brazil}

\usepackage{algorithm}
\usepackage{algpseudocode}
\usepackage{pifont}
\usepackage{booktabs}
\usepackage{multirow}
\usepackage{xcolor}
\usepackage{placeins}
\usepackage{graphicx}
\begin{document}

\title{COMEX: A Composition-Grounded Benchmark and Learning Framework for Explainable Aesthetic Image Cropping}



\author{Rui Yang}
\affiliation{
  \institution{State Key Laboratory of Mobile Network and Mobile Multimedia Technology, ZTE Corporation}
  \city{Shenzhen}
  \country{China}}
  \orcid{0000-0002-3209-0456}
\email{yang.rui15@zte.com.cn}

\author{Wei Zhou}
\authornote{Equal Contribution}
\affiliation{
  \institution{State Key Laboratory of Mobile Network and Mobile Multimedia Technology, ZTE Corporation}
  \city{Shenzhen}
  \country{China}}
  \orcid{/0000-0002-5794-7567}
\email{zhou.wei37@zte.com.cn}

\author{Dingyong Gou}
\authornote{Corresponding Author}
\affiliation{
  \institution{State Key Laboratory of Mobile Network and Mobile Multimedia Technology, ZTE Corporation}
  \city{Shenzhen}
  \country{China}}
  \orcid{/0009-0003-1261-5555}
\email{gou.dingyong@zte.com.cn}

\author{Xiaohui Cui}
\affiliation{
  \institution{State Key Laboratory of Mobile Network and Mobile Multimedia Technology, ZTE Corporation}
  \city{Shenzhen}
  \country{China}}
  \orcid{/0009-0007-9332-0685}
\email{cui.xiaohui1@zte.com.cn}

\author{Cong Li}
\affiliation{
  \institution{State Key Laboratory of Mobile Network and Mobile Multimedia Technology, ZTE Corporation}
  \city{Shenzhen}
  \country{China}}
  \orcid{/0009-0003-0783-0664}
\email{li.cong4@zte.com.cn}

\author{Yinyin Gong}
\affiliation{
  \institution{State Key Laboratory of Mobile Network and Mobile Multimedia Technology, ZTE Corporation}
  \city{Shenzhen}
  \country{China}}
  \orcid{/0009-0009-9349-7901}
\email{gong.yinyin1@zte.com.cn}

\author{Yipo Huang}
\affiliation{
  \institution{School of Data Science and Institute of Artificial Intelligence, Chang'an University}
  \city{Xi'an}
  \country{China}}
  \orcid{/0000-0003-0908-2180}
\email{yphuang@chd.edu.cn}

\author{Jiliang Zhao}
\affiliation{
  \institution{State Key Laboratory of Mobile Network and Mobile Multimedia Technology, ZTE Corporation}
  \city{Shenzhen}
  \country{China}}
  \orcid{/0009-0002-6009-5323}
\email{zhao.jiliang@zte.com.cn}








\renewcommand{\shortauthors}{Trovato et al.}

\begin{abstract}
Explainable aesthetic image cropping requires not only localizing a visually pleasing crop but also explaining why it is preferred. Existing crop-and-explain methods largely treat explanation as post-hoc text generation and overlook composition, a key aesthetic factor that links crop decisions with interpretable reasoning. In this paper, we reformulate explainable aesthetic image cropping as a structured crop-composition-explanation problem. To support this setting, we introduce COMEX, a new benchmark built through image expansion and an IO-reversal pipeline. COMEX contains 33,161 quadruples, each consisting of an expanded image, a crop box, a composition category, and a composition-grounded explanation, enabling joint learning of crop localization, composition understanding, and explanation generation. We further propose a two-stage SFT+GRPO framework, where supervised fine-tuning establishes the structured output protocol and basic cropping ability, and GRPO further improves crop quality, composition prediction, and explanation faithfulness. We benchmark 15 large vision-language models and existing cropping methods on COMEX, establishing a comprehensive testbed for composition-grounded explainable aesthetic cropping. Experiments on both COMEX and prior benchmarks demonstrate the effectiveness and transferability of our framework, with strong performance across evaluation metrics.
\end{abstract}




\maketitle



\section{Introduction}

\begin{figure*}
  \centering
  \includegraphics[width=0.85\textwidth]{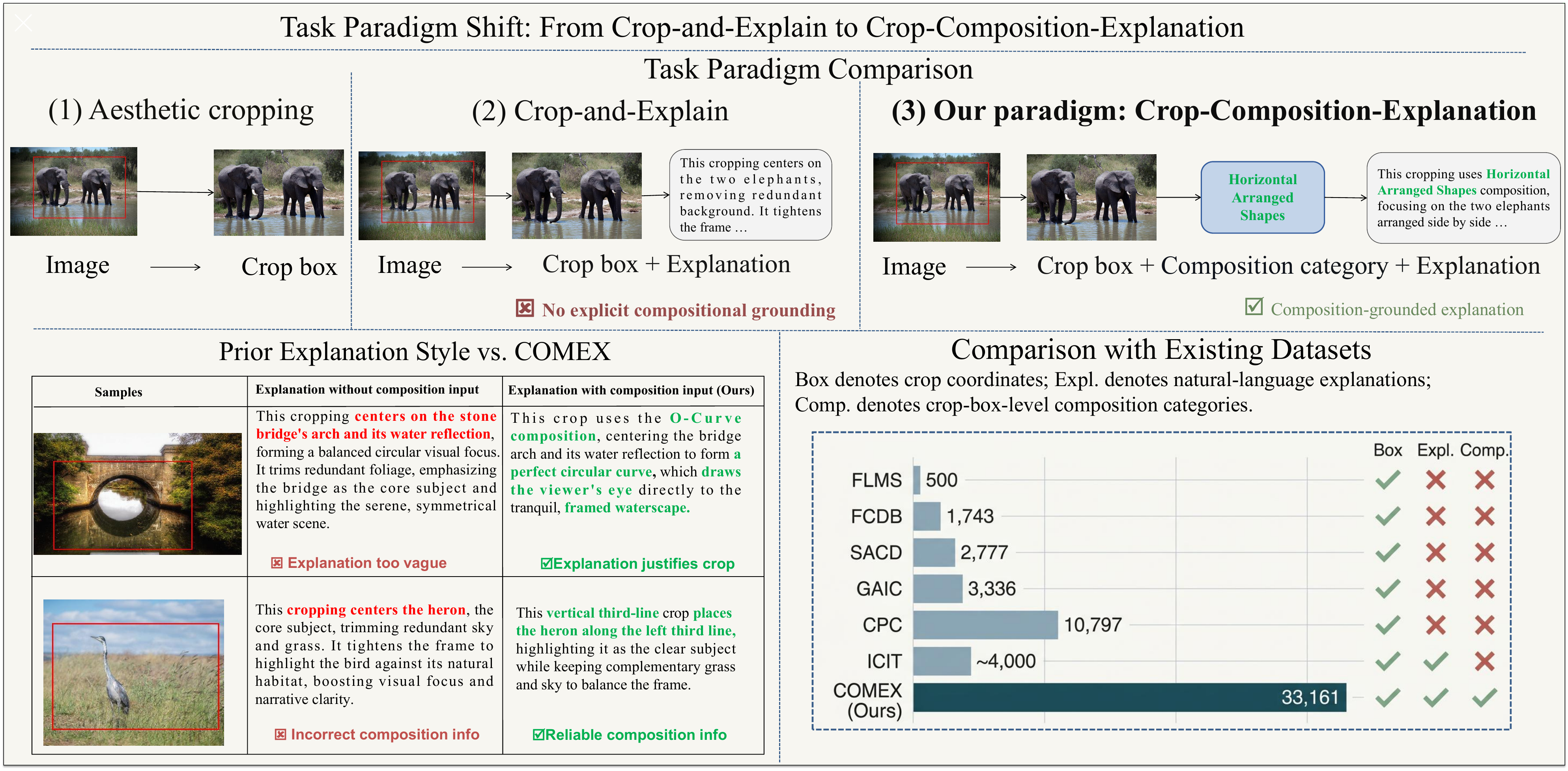}
  \caption{COMEX vs. Prior Explanation Methods.}
  \Description{A comparison figure showing the difference between prior crop-and-explain methods and COMEX in terms of task paradigm and data format.}
  \label{datashow}
\end{figure*}
Aesthetic image cropping aims to predict a visually pleasing crop box from an input image, with broad applications in mobile photography, photo post-processing, and intelligent content editing \cite{sheng2025instructcrop,hong2024learning,hong2021composing,guo2018automatic}. Unlike object detection \cite{redmon2016you} or saliency prediction \cite{wang2018detect}, it requires not only preserving the main subject and semantic integrity, but also accounting for compositional principles, spatial layout, and aesthetic expressiveness \cite{ni2013learning}. The task therefore lies at the intersection of geometric localization and high-level aesthetic reasoning. 
Early methods mainly relied on candidate-window ranking~\cite{wei2018good,zeng2020grid,zeng2019reliable,tu2020image} or direct crop-box regression~\cite{pan2021robust}. However, most of these approaches focus on predicting \emph{where to crop}, while offering limited insight into \emph{why} a particular crop is aesthetically preferable.
With the rise of multimodal large language models \cite{bai2025qwen3,liu2023visual,du2026venus,huang2024aesexpert}, recent explainable cropping
methods~\cite{sheng2025instructcrop,du2026venus} introduce a
unified \emph{crop-and-explain} setting, moving aesthetic image
cropping beyond pure box prediction toward explainable generation.

Despite this progress, existing \emph{crop-and-explain}
methods~\cite{sheng2025instructcrop} do not explicitly model
composition. As a fundamental principle in photography, composition governs subject emphasis, spatial arrangement, and visual balance~\cite{hong2021composing}. In practice, photographers make cropping decisions largely based on compositional rules such as the rule of thirds and perspective~\cite{ni2013learning}. These principles not only determine \emph{where} a crop should be placed, but also provide a concise and interpretable rationale for \emph{why} that placement is aesthetically desirable. Without this intermediate reasoning layer, models can only infer explanations retrospectively from the predicted crop box, often resulting in descriptions that are vague, generic, or misaligned with the underlying aesthetic intent. We therefore
propose to formulate the task as a structured
\emph{crop--composition--explanation} problem, where composition
naturally bridges crop boxes and their explanations. As illustrated in
Figure~\ref{datashow} (lower left), introducing composition as an
explicit intermediate yields explanations that are more specific,
aesthetically grounded, and faithful to the underlying cropping
rationale.

Realizing this formulation requires advances in both benchmark
construction and training methodology. On the data side, current
datasets~\cite{sheng2025instructcrop,du2026venus} are typically built
by augmenting existing cropping datasets with large-model-generated
explanations followed by manual refinement. As a result, they are
limited in scale and, more critically, lack composition annotations
explicitly aligned with crop boxes, often yielding generic
explanations only weakly grounded in cropping rationale. To address
this, we introduce the \textbf{COM}position-grounded aesthetic
cropping-and-\textbf{EX}planation dataset (\textbf{COMEX}), built in
two main stages on the large-scale composition classification dataset
PICD~\cite{zhao2025can}: first, image expansion and an input--output
reversal (IO-reversal) procedure transform image--composition pairs
into triples of expanded images, crop boxes, and composition
categories; second, under composition-category constraints, multimodal
large models generate compositionally grounded explanations for each
crop. In total, COMEX contains \textbf{33,161} quadruples of expanded
images, crop boxes, composition categories, and composition-grounded
explanations, making it the first large-scale benchmark for
explainable aesthetic cropping to jointly provide all three types of
annotation. A comparison with existing
datasets is shown in Figure~\ref{datashow} (lower right).

On the methodology side, existing approaches rely solely on supervised
fine-tuning (SFT), learning to generate explanations by imitating
reference texts without explicit composition guidance. While SFT can
produce reasonable outputs, it optimizes only token-level likelihood
and provides no direct feedback on cropping quality or explanation
faithfulness, limiting further improvement in composition-grounded
reasoning. To overcome this limitation, we propose a two-stage
\textbf{SFT\,+\,GRPO} framework. The first stage uses structured SFT
to teach the model to produce crop boxes, composition categories, and
explanations in a unified format. Empirically, extending SFT beyond
convergence yields only marginal gains across all metrics
(Table~\ref{tab:stage2_necessity}), confirming the inherent limitation
of imitation learning. The second stage therefore introduces Group
Relative Policy Optimization (GRPO)~\cite{guo2025deepseek}, replacing
token-level imitation with task-aligned rewards targeting cropping
quality, composition consistency, and explanation quality. By directly
optimizing these decision-level objectives, GRPO brings consistent
improvements beyond continued SFT---notably in crop precision,
composition prediction, and explanation specificity---without requiring
an additional value model.

Experiments confirm that composition supervision is crucial for
explainable aesthetic cropping, improving both IoU and explanation
quality, and that GRPO-based reinforcement learning further
strengthens the Stage~I SFT baseline in both aspects. We benchmark
COMEX against existing methods and 15 large vision-language models,
establishing a comprehensive evaluation baseline. Strong transfer
results on prior benchmarks further demonstrate the generalizability
of both COMEX and our framework, which consistently outperforms
compared models across multiple evaluation dimensions.

Our contributions are three-fold:

\begin{itemize}
    \item We reformulate explainable aesthetic image cropping as a \emph{crop--composition--explanation} task, explicitly modeling composition as the intermediate layer between crop decision and explanation generation.
    
    \item We build \textbf{COMEX}, the first large-scale benchmark for this task, containing 33,161 samples with aligned annotations of crop boxes, composition categories, and composition-grounded explanations.
    
    \item We propose a two-stage \textbf{SFT+GRPO} framework that jointly improves crop localization, composition prediction, and explanation generation. Extensive experiments, including ablations and transfer evaluations, demonstrate its effectiveness and generalizability over existing vision-language baselines.
\end{itemize}

\begin{figure*}[!t]
  \centering
   \includegraphics[width=0.85\linewidth]{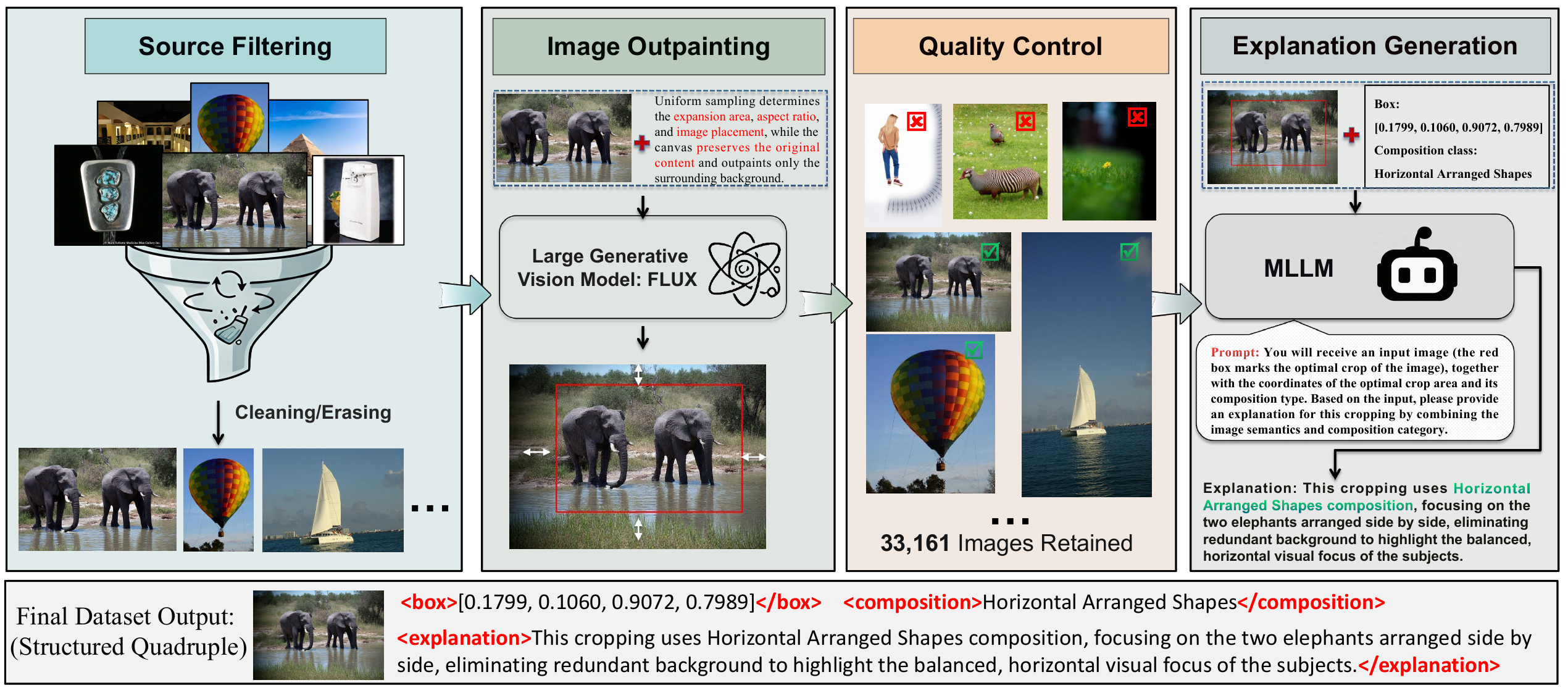} 
   \caption{The pipeline for constructing the proposed COMEX}
   \label{data pipeline}
\end{figure*}

\section{Related Work}

\subsection{Aesthetic Image Cropping}

Aesthetic image cropping aims to predict a visually pleasing crop that preserves the main subject while improving composition and overall visual quality. Existing methods can be broadly grouped into two categories: candidate-scoring~\cite{chen2017learning,Wei_2018_CVPR,zeng2020grid} and direct regression~\cite{guo2018automatic,li2018a2,hong2021composing}.
\textbf{Candidate-scoring methods} first generate multiple crop proposals and then rank them by aesthetic quality. Early studies formulate cropping as a view-ranking problem learned from professional photographs~\cite{chen2017learning} or dense comparative crop pairs~\cite{Wei_2018_CVPR}. Subsequent works improve efficiency and reliability by combining attention prediction with aesthetic ranking~\cite{wang2018deep}, reducing the search space with grid anchors and stronger benchmarks~\cite{zeng2020grid}, or introducing composition- and saliency-aware score maps for more interpretable selection~\cite{tu2020image}.
\textbf{Direct regression methods} instead predict crop coordinates directly, avoiding explicit candidate enumeration. Representative examples include cascaded regression based on deep aesthetic features~\cite{guo2018automatic}, reinforcement-learning-based sequential cropping~\cite{li2018a2}, and later methods that incorporate stronger composition cues, such as salient-cluster-guided cropping~\cite{pan2021robust} and key-composition-map-guided regression~\cite{hong2021composing}. More recent extensions further explore richer supervision and controllability, including subject-aware cropping from pseudo pairs~\cite{hong2024learning} and intent-aware cropping with vision-language conditioning~\cite{zhong2023clipcrop}.
Most methods predict \emph{where to crop} but not \emph{why}.
Although InstructCrop~\cite{sheng2025instructcrop} jointly generates
crops and explanations, its explanations remain weakly grounded in
composition principles. We explicitly model composition and formulate
cropping as a structured
\emph{crop--composition--explanation} problem.


\subsection{Reinforcement Learning and GRPO for Vision-Language Models}

Reinforcement learning has become an important post-training strategy
for improving large-model reasoning and
generation~\cite{wang2022deep,guo2025deepseek,plaat2022deep,comanici2025gemini,yang2025qwen3,team2025kimi}.
Among these methods, GRPO, introduced in
DeepSeek-R1~\cite{guo2025deepseek}, optimizes relative rewards within
grouped samples without requiring a separate value model, offering a
practical way to improve decision quality beyond reference-text
imitation. This paradigm has been extended to multimodal settings:
Visual-RFT~\cite{liu2025visual} applies GRPO-style training to visual
perception tasks such as detection and grounding,
MedVLM-R1~\cite{pan2025medvlm} shows that reinforcement learning can
elicit explicit reasoning in medical vision-language models without
reasoning annotations, and Q-Insight~\cite{li2025q} demonstrates
gains in visual quality understanding and comparative reasoning. Unlike standard GRPO tasks with directly verifiable rewards,
explainable cropping jointly optimizes crop quality, composition
consistency, and explanation faithfulness. Our \textbf{SFT+GRPO}
framework first learns structured crops, composition categories, and
explanations, and then refines them through reward optimization,
aligning with our formulation and COMEX supervision.


\section{The COMEX Benchmark}
Existing aesthetic image cropping
datasets~\cite{chen2017learning,celona2019autocropping,yang2023focusing,sheng2025instructcrop,zeng2020grid,zhang2022human}
either lack language explanations altogether or provide only post-hoc
textual descriptions without explicit composition annotations tied to
crop boxes (Figure~\ref{datashow}). To address this gap, we construct
\textbf{COMEX} (\textbf{COM}position-guided aesthetic cropping-and-%
\textbf{EX}planation), a large-scale benchmark that provides
structured quadruples of \textit{(expanded image, crop box,
composition category, explanation)}, jointly supporting crop
localization, composition recognition, and composition-grounded
explanation generation under a unified evaluation protocol.

\subsection{Data Source and IO-Reversal}

We build upon PICD~\cite{zhao2025can}, a large-scale composition
classification dataset containing 49,123 expert-annotated images
spanning 24 composition categories. The images are sourced from
professional photography websites~\cite{flickr,pexels,unsplash}
and curated aesthetic datasets~\cite{murray2012ava,
kuznetsova2020open,yang2022personalized}.The diverse sourcing ensures high aesthetic
quality and broad compositional coverage, and each image carries a
professional composition label, making PICD a reliable foundation for
composition-aware cropping.

Rather than manually collecting crop boxes or relying on
model-predicted crops, we adopt an \textit{IO-reversal} strategy:
each original PICD image is treated as the ideal crop target, and a
larger context image is constructed through outpainting. The crop box
thus naturally inherits the composition label from the original
annotation, providing crop-level composition supervision without
additional relabeling. A key property of this design is that the
original photograph is preserved pixel-for-pixel inside the expanded
canvas: the crop target seen during training is always a real image,
and only the surrounding context is synthetically generated. Any
synthetic-to-real domain gap is therefore confined to the contextual
background rather than the aesthetic content the model needs to learn
from.

\subsection{Construction Pipeline}

The dataset construction proceeds through four sequential stages, as
illustrated in Figure~\ref{data pipeline}.

\noindent\textbf{(1) Source Filtering.}
We begin by cleaning the raw PICD images, removing samples with
excessive overlaid text, watermarks, large black borders, or other
quality defects, as such artifacts would degrade the crop target and
compromise composition annotations. After filtering, 44,201 images
are retained.

\noindent\textbf{(2) Image Outpainting.}
Each filtered image is expanded using
FLUX.1~[dev]~\cite{labs2025flux1kontextflowmatching}, a
state-of-the-art text-guided generative model. The expansions cover
area ratios from $1\times$ to $10\times$ the original image size,
with aspect ratios ranging from standard mobile portrait formats to
ultra-wide panoramic layouts. Since the outpainted region serves
solely as surrounding context from which the model must recover the
original crop, the expansion should appear natural and avoid
introducing extraneous subjects. To this end, we employ a carefully
designed prompt that promotes realism and suppresses common generation
artifacts: \textit{``natural and realistic, seamless integration, no
background blur, no borders, no stitching artifacts, expansive
scenes, no obstructions, no additional subjects.''}

\noindent\textbf{(3) Quality Control.}
Despite the use of guided prompts, generative outpainting can still
introduce visual defects in the surrounding context. We therefore
conduct a manual inspection stage, discarding samples whose expanded
regions exhibit unnatural generation artifacts, physically implausible
structures, or non-photographic layouts (e.g., poster- or
webpage-style images), removing approximately 25\% of the outpainted
samples. This ensures that the retained images appear natural and
free of noticeable generation or outpainting traces, making them
suitable for downstream cropping.

\noindent\textbf{(4) Explanation Generation.}
For each sample that passes visual inspection, we generate a
composition-grounded cropping explanation using Seed-1.8~\cite{seed20_bytedance}, a multimodal large language
model. The model receives the expanded image with the crop region
highlighted by a prominent red bounding box, along with the crop box
coordinates and the associated composition category. It is instructed
to interpret the crop in terms of both subject semantics and
compositional principles, ensuring that the generated explanations are
grounded in crop-level composition annotations rather than serving as
generic post-hoc descriptions. We then perform a second round of
manual review to discard explanations that are factually inconsistent
with the image content or fail to reference the annotated composition
category. After both rounds of quality control, 33,161 valid samples
are retained for the final benchmark.

\subsection{Comparison with Existing Datasets}

As shown in Figure~\ref{datashow}, COMEX is the largest benchmark for
explainable aesthetic image cropping, containing 33,161 annotated
samples---more than three times the size of the previous largest
dataset, CPC~\cite{wei2018good} (10,797). Existing datasets mainly
provide crop box annotations; ICIT~\cite{sheng2025instructcrop} is
the only prior benchmark that additionally includes natural-language
explanations, yet none of them offers crop-box-level composition
annotations. COMEX fills this gap by jointly annotating crop boxes,
composition categories, and composition-grounded explanations. 


The qualitative comparison in Figure~\ref{datashow} (lower right)
further illustrates the value of composition supervision. Without
composition information, generated explanations tend to be vague or
inconsistent with the actual crop rationale, whereas COMEX enables
more precise and faithful explanations by explicitly grounding them in
composition categories. Notably, these composition labels are
inherited from expert annotations in PICD rather than predicted by
models, ensuring high annotation reliability. This unified annotation
design allows models to learn not only \emph{where} to crop, but also
\emph{under which compositional principle} and \emph{why}.


\begin{figure*}[h]
  \centering
   \includegraphics[width=0.85\linewidth]{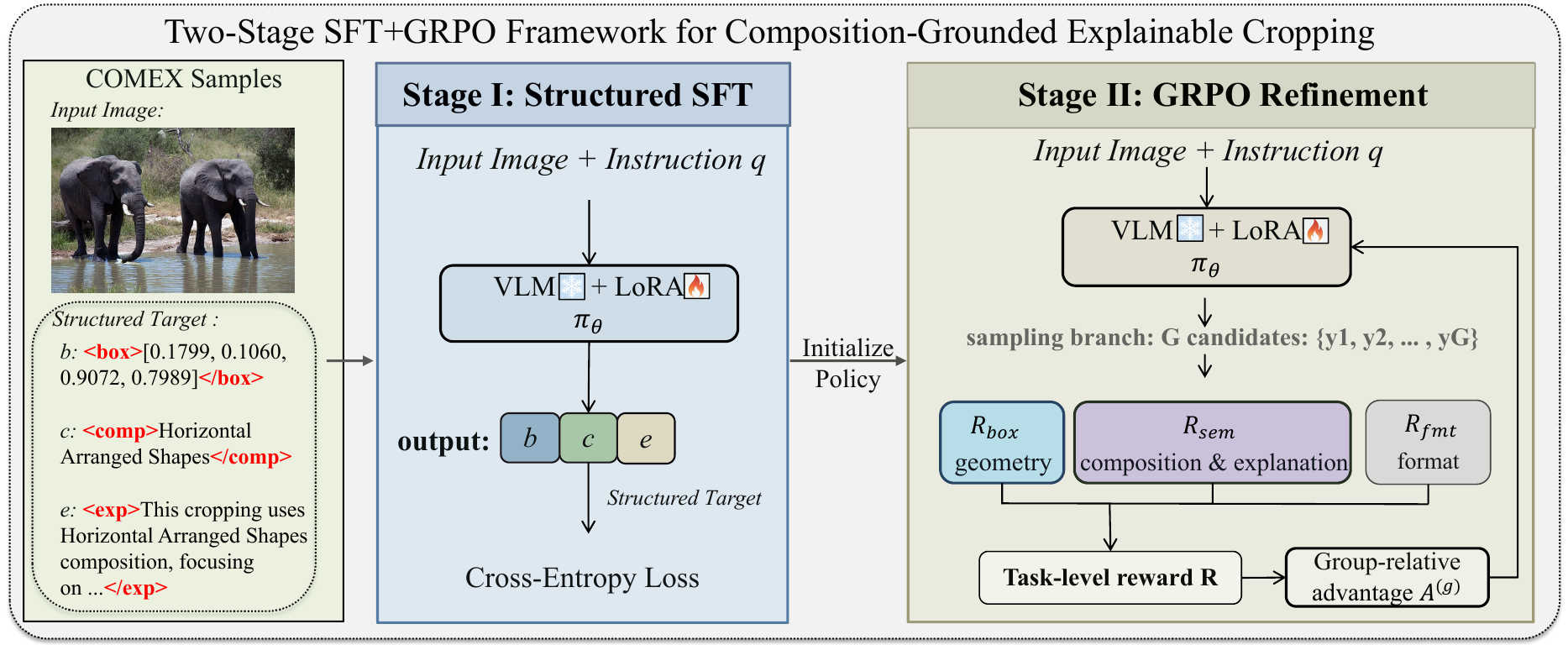} 
   \caption{Overview of two-stage SFT+GRPO framework for explainable aesthetic image cropping.}
   \label{method pipeline}
\end{figure*}

\section{Method}
\subsection{Problem Formulation}

We formulate explainable aesthetic image cropping as a conditional
structured generation problem. Given an input image $I$ and a textual
instruction $q$ that specifies the cropping task, the model
generates a structured triplet
\begin{equation}
    y = (b,\; c,\; e),
\end{equation}
where $b=(x_1,y_1,x_2,y_2)$ denotes the crop box coordinates, $c$
is the composition category, and $e$ is a natural-language explanation
grounding the crop decision. The three components are generated
sequentially in a unified textual format compatible with
vision-language models (VLMs). Let the training set be

\begin{equation}
    \mathcal{D}=\{(I_i,\,b_i,\,c_i,\,e_i)\}_{i=1}^{N}.
\end{equation}

Our goal is to learn a policy $\pi_\theta(y\mid I,q)$ that jointly
produces geometrically accurate crop boxes, composition-consistent
category labels, and decision-grounded natural-language explanations.

\begin{table}[t]
\centering
\small
\setlength{\tabcolsep}{4pt}
\renewcommand{\arraystretch}{1.15}
\caption{Main results on COMEX. We compare classic aesthetic cropping
methods, MLLM-based cropping methods, and zero-shot general-purpose
MLLMs against our two-stage framework. ``Comp-ACC'': composition
accuracy; ``*'': our reproduction; ``--'': unavailable. Best in
\textbf{bold}.}
\label{tab:model_results_grouped}
\resizebox{\columnwidth}{!}{%
\begin{tabular}{l c c c c}
\toprule
\textbf{Model} & \textbf{mIoU}$\uparrow$ & \textbf{mBDE}$\downarrow$ & \textbf{Comp-ACC}$\uparrow$ & \textbf{METEOR}$\uparrow$ \\
\midrule
\multicolumn{5}{l}{\textit{Classic Aesthetic Cropping}} \\
\midrule
CACNet~\cite{hong2021composing}\,{\scriptsize CVPR'21} & 0.7603 & 0.0456 & -- & -- \\
AR2L~\cite{li2018a2}\,{\scriptsize CVPR'18} & 0.6304 & 0.0858 & -- & -- \\
MARS~\cite{li2020learning}\,{\scriptsize CVPR'20} & 0.4370 & 0.2386 & -- & -- \\
\midrule
\multicolumn{5}{l}{\textit{MLLM-based Aesthetic Cropping}} \\
\midrule
Venus (ZS)~\cite{du2026venus}\,{\scriptsize CVPR'26} & 0.5891 & 0.1012 & -- & -- \\
Venus~\cite{du2026venus}\,{\scriptsize CVPR'26} & 0.7102 & 0.0611 & -- & -- \\
InstructCrop (ZS)~\cite{sheng2025instructcrop}\,{\scriptsize MM'25} & 0.5970 & 0.0759 & -- & -- \\
InstructCrop*~\cite{sheng2025instructcrop}\,{\scriptsize MM'25} & 0.6318 & 0.0616 & -- & -- \\
\midrule
\multicolumn{5}{l}{\textit{General-purpose MLLMs (zero-shot)}} \\
\midrule
Qwen3.5 9B~\cite{qwen35_hf_collection} & 0.4895 & 0.1281 & -- & 0.2708 \\
Qwen3.5 4B~\cite{qwen35_hf_collection} & 0.4817 & 0.1326 & -- & 0.2443 \\
Qwen3.5 2B~\cite{qwen35_hf_collection} & 0.4774 & 0.1383 & -- & 0.1870 \\
InternVL3.5-1B-Flash~\cite{wang2025internvl3} & 0.2000 & 0.1826 & -- & 0.2163 \\
InternVL3.5-8B~\cite{wang2025internvl3} & 0.3564 & 0.1731 & -- & 0.1389 \\
MiniMax-M2.5~\cite{minimax_m25_hf} & 0.5064 & 0.1249 & -- & 0.2252 \\
Qwen3-VL-32B~\cite{bai2025qwen3} & 0.5344 & 0.1137 & -- & 0.2562 \\
Qwen3-VL-8B~\cite{bai2025qwen3} & 0.4854 & 0.1295 & -- & 0.2384 \\
Qwen3-VL-2B~\cite{bai2025qwen3} & 0.5394 & 0.1267 & -- & 0.1337 \\
GLM-4.6V-Flash~\cite{glm46_hf} & 0.5081 & 0.1296 & -- & 0.0695 \\
Seed2.0 Pro~\cite{seed20_bytedance} & 0.5071 & 0.1247 & -- & 0.2245 \\
Seed1.8~\cite{seed20_bytedance} & 0.5097 & 0.1242 & -- & 0.2233 \\
Seed1.6~\cite{seed20_bytedance} & 0.5087 & 0.1245 & -- & 0.2242 \\
Step3-VL-10B~\cite{step3vl10b_hf} & 0.4917 & 0.1279 & -- & 0.1113 \\
MiniCPM-V-4.5~\cite{minicpmv4_hf} & 0.4670 & 0.1362 & -- & 0.2114 \\
\midrule
\multicolumn{5}{l}{\textit{Ours}} \\
\midrule
Qwen3.5-2B (SFT) & 0.7105 & 0.0611 & 0.7120 & 0.4994 \\
Qwen3.5-0.8B (SFT) & 0.6927 & 0.0661 & 0.6903 & 0.4848 \\
Qwen3.5-0.8B (SFT+GRPO) & 0.7015 & 0.0640 & 0.6914 & 0.4875 \\
Qwen3-VL-2B (SFT) & 0.7532 & 0.0509 & 0.7404 & 0.5129 \\
\textbf{Qwen3-VL-2B (SFT+GRPO)} & \textbf{0.7765} & \textbf{0.0447} & \textbf{0.7448} & \textbf{0.5194} \\
\bottomrule
\end{tabular}%
}
\end{table}

\subsection{Two-Stage SFT+GRPO Framework}

We adopt a two-stage training framework: the first stage establishes
reliable structured generation via supervised fine-tuning, while the
second stage refines task-level decision quality through reinforcement
learning. An overview is illustrated in Figure~\ref{method pipeline}.

\paragraph{Stage~I: Structured Supervised Fine-Tuning.}
We train the model with standard autoregressive cross-entropy loss
over the structured target sequence $(b, c, e)$:
\begin{equation}
    \mathcal{L}_{\mathrm{SFT}}
    = -\sum_{t=1}^{T}\log \pi_\theta(y_t\mid y_{<t},\,I,\,q),
\end{equation}
where $y_t$ denotes the $t$-th target token. This stage yields an
initial policy $\pi_{\theta_0}$ that reliably generates well-formed
crop boxes, composition categories, and explanations in the required
output format, providing a solid initialization for Stage~II and
preventing the reward hacking that can arise when applying RL to
structured generation from scratch.

\paragraph{Stage~II: GRPO-based Reinforcement Learning.}
Starting from $\pi_{\theta_0}$, we further optimize the model with
Group Relative Policy Optimization (GRPO)~\cite{guo2025deepseek}.
For each input $(I, q)$, the current policy samples a group of $G$
candidate outputs:
\begin{equation}
    \{y^{(g)}\}_{g=1}^{G},
    \qquad y^{(g)} \sim \pi_\theta(\cdot \mid I,\,q).
\end{equation}
Each candidate is assigned a task-level reward $R^{(g)}$, and the
policy is updated via the group-relative advantage:
\begin{equation}
    A^{(g)}=\frac{R^{(g)}-\mu_R}{\sigma_R+\varepsilon},
\end{equation}
where $\mu_R$ and $\sigma_R$ are the within-group mean and standard
deviation of rewards, and $\varepsilon > 0$ is a small stabilizer.
By estimating advantages through relative ranking within the sampled
group, GRPO eliminates the need for a separate value network~\cite{guo2025deepseek},
substantially reducing training overhead while enabling Stage~II to
improve crop quality and explanation faithfulness beyond what
supervised imitation alone can achieve.

We implement our framework on an open-source vision-language backbone
and adopt LoRA~\cite{hulora} for parameter-efficient tuning in
both Stage~I SFT and Stage~II GRPO. All stages share the same prompt
template and structured output format. Full prompt details are
provided in the supplementary material.

\subsection{Reward Design}

The Stage-II reward consists of three complementary terms that
jointly supervise the output: a geometric reward
$R_{\mathrm{box}}$ for crop spatial accuracy, a semantic explanation
reward $R_{\mathrm{sem}}$ for explanation faithfulness and
compositional consistency, and a format term
$R_{\mathrm{fmt}}$ for output structure and language quality.

\paragraph{Geometric reward.}
$R_{\mathrm{box}}$ combines region overlap with boundary precision:
\begin{equation}
R_{\mathrm{box}}
  = (1 - w_{\mathrm{bde}})\, R_{\mathrm{iou}}
  + w_{\mathrm{bde}}\, R_{\mathrm{bde}},
\end{equation}
where $R_{\mathrm{iou}}$ rewards high Intersection-over-Union between
the predicted crop box and the ground truth, and $R_{\mathrm{bde}}$
penalizes boundary displacement error to encourage precise edge
alignment. We set $w_{\mathrm{bde}} = 0.25$.

\paragraph{Semantic explanation reward.}
$R_{\mathrm{sem}}$ enforces two forms of consistency:
\begin{equation}
R_{\mathrm{sem}}
  = \lambda_s\, R_{\mathrm{exp}}
  + (1 - \lambda_s)\, R_{\mathrm{comp}},
\end{equation}
where $R_{\mathrm{exp}}$ is an explanation reward that measures the
similarity between the generated explanation and the ground-truth
reference, ensuring linguistic fidelity, and $R_{\mathrm{comp}}$ is a
composition reward that evaluates whether the explanation is
semantically aligned with the ground-truth composition category,
encouraging compositional grounding. We set $\lambda_s = 0.7$.

\paragraph{Format term.}
Following prior work on reinforcement learning with structured
outputs~\cite{li2025q}, we also introduce a format reward
$R_{\mathrm{fmt}}$ that constrains valid output syntax and promotes
well-formed explanations via length adequacy and semantic coverage.

\paragraph{Overall reward.}
The total reward aggregates all three terms as:
\begin{equation}
R_{\mathrm{all}}
  = \alpha_g\, R_{\mathrm{box}}
  + \alpha_s\, R_{\mathrm{sem}}
  + \alpha_f\, R_{\mathrm{fmt}},
\end{equation}
where $\alpha_g$, $\alpha_s$, and $\alpha_f$ weight the geometric, semantic, and format rewards, respectively. We set them to $0.5$, $0.175$, and $0.325$ based on validation performance. This design enables GRPO to jointly optimize spatial, semantic, and structural quality. Detailed definitions of all sub-reward terms are provided in the supplementary material.

\section{Experiments}

\subsection{Experimental Setup}

\paragraph{Datasets and Benchmarks.}
We use \textbf{COMEX} as the primary benchmark for explainable
aesthetic image cropping, with 26,528 images for training and 6,633
for testing. To evaluate cross-dataset generalization, we also report
results on \textbf{FCDB}~\cite{chen2017quantitative} (1,743 images;
1,395 for training, 348 for testing, following prior work).

\paragraph{Evaluation Metrics.}
We evaluate three aspects of performance. Crop quality is measured by
\textbf{IoU} (overlap between predicted and ground-truth crops) and
\textbf{BDE} (boundary displacement error), following
InstructCrop~\cite{sheng2025instructcrop}. Composition prediction is
measured by \textbf{Comp-ACC}, the accuracy of the predicted
composition category against the ground-truth label. Explanation
quality is measured by \textbf{METEOR}~\cite{banerjee2005meteor},
which captures the similarity between generated and reference
explanations. Together, these metrics cover geometric accuracy,
composition correctness, and explanation faithfulness.

\subsection{Main Results}

\subsubsection{Results on COMEX}

Table~\ref{tab:model_results_grouped} summarizes the main results on
COMEX. Our method consistently outperforms general-purpose MLLMs,
prior MLLM-based cropping approaches, and classic aesthetic cropping
baselines across all evaluated metrics.

Under zero-shot prompting, all 15 general-purpose MLLMs struggle with
crop localization. The best zero-shot model, Qwen3-VL-2B, reaches
only 0.5394 mIoU, while larger variants such as Qwen3-VL-32B achieve
0.5344, indicating that scaling alone does not resolve the task.
Among prior MLLM-based cropping methods, the strongest, Venus,
achieves 0.7102 mIoU and 0.0611 mBDE. Our Qwen3-VL-2B with Stage~I
SFT already surpasses it with 0.7532 mIoU and 0.0509 mBDE, and
Stage~II widens the gap to 0.7765 and 0.0447. Our final model also
outperforms the strongest classic baseline, CACNet, in both mIoU
(0.7765 vs.\ 0.7603) and mBDE (0.0447 vs.\ 0.0456). Unlike all
prior cropping methods, ours additionally provides composition
classification and natural-language explanations.

Our framework also generalizes across backbones. With the smaller
Qwen3.5-0.8B, Stage~I SFT already achieves 0.6927 mIoU, and
Stage~II further improves it to 0.7015. With Qwen3.5-2B, Stage~I
reaches 0.7105 mIoU, matching Venus while also producing
composition and explanation outputs. These results confirm that the
proposed two-stage framework is effective regardless of backbone
choice.

Stage~II consistently improves all metrics over Stage~I: on
Qwen3-VL-2B, mIoU rises from 0.7532 to 0.7765, mBDE drops from
0.0509 to 0.0447, composition accuracy improves from 0.7404 to
0.7448, and METEOR from 0.5129 to 0.5194. This confirms that the
reinforcement learning stage benefits crop localization, composition
understanding, and explanation quality alike. A detailed analysis is
provided in the ablation studies.

\begin{table}[t]
\centering
\small
\setlength{\tabcolsep}{6pt}
\renewcommand{\arraystretch}{1.12}
\caption{Comparison on FCDB (IoU / BDE). ``COMEX, zero-shot transfer'' denotes
models trained on COMEX and evaluated on FCDB without fine-tuning.
Best in \textbf{bold}.}

\label{tab:fcdb_results}
\resizebox{\columnwidth}{!}{
\begin{tabular}{l l c c}
\toprule
\textbf{Method} & \textbf{Publication} & \textbf{IoU} $\uparrow$ & \textbf{BDE} $\downarrow$ \\
\midrule

\multicolumn{4}{l}{\textit{Classic / prior cropping methods}} \\
\midrule
VFN~\cite{chen2017learning}         & ACM MM 2017 & 0.6850 & 0.0840 \\
DIC~\cite{wang2017deep}             & ICCV 2017   & 0.6300 & 0.0900 \\
DIC*~\cite{wang2018deep}            & TPAMI 2018  & 0.6500 & 0.0800 \\
A2-RL~\cite{li2018a2}               & CVPR 2018   & 0.6640 & 0.0890 \\
Fast A3-RL~\cite{li2019fast}        & TIP 2019    & 0.6960 & 0.0770 \\
CGS~\cite{li2020composing}          & CVPR 2020   & --     & --     \\
CACNet~\cite{hong2021composing}     & CVPR 2021   & 0.7020 & 0.0740 \\
GAIC~\cite{zeng2020grid}            & TPAMI 2022  & 0.6740 & 0.0810 \\
Jia et al.~\cite{jia2022rethinking} & CVPR 2022   & --     & --     \\
UNIC~\cite{liu2023beyond}           & ICCV 2023   & --     & --     \\
Wang et al.~\cite{wang2023image}    & CVPR 2023   & 0.6950 & 0.0750 \\
Cropper~\cite{lee2025cropper}       & CVPR 2025   & 0.6670 & 0.0870 \\
\midrule

\multicolumn{4}{l}{\textit{MLLM-based cropping methods}} \\
\midrule
InstructCrop~\cite{sheng2025instructcrop}       & ACM MM 2025 & 0.7100 & 0.0720 \\
InstructCrop (exp)~\cite{sheng2025instructcrop} & ACM MM 2025 & 0.7160 & 0.0700 \\
\midrule

\multicolumn{4}{l}{\textit{General-purpose MLLMs}} \\
\midrule
GPT-4o~\cite{hurst2024gpt}                      & -- & 0.2170 & 0.3150 \\
Qwen2.5-VL-max~\cite{bai2025qwen25vltechnicalreport} & -- & 0.1290 & 0.3850 \\
Qwen3-VL-2B~\cite{bai2025qwen3}                 & -- & 0.3959 & 0.4048 \\
\midrule

\multicolumn{4}{l}{\textit{Ours (COMEX, zero-shot transfer)}} \\
\midrule
Qwen3-VL-2B Stage~I        & -- & 0.5946 & 0.1035 \\
Qwen3-VL-2B Stage~I+II     & -- & 0.6251 & 0.0938 \\
\midrule

\multicolumn{4}{l}{\textit{Ours (trained on FCDB)}} \\
\midrule
Qwen3-VL-2B Stage~I            & -- & 0.7007 & 0.0743 \\
\textbf{Qwen3-VL-2B Stage~I+II} & -- & \textbf{0.7225} & \textbf{0.0676} \\
\bottomrule
\end{tabular}
}
\end{table}

\begin{table}[h]
\centering
\small
\caption{Effect of composition information in Stage~I SFT (6,633-image test set).
Explanation quality is measured by pairwise win rate (\%) from three MLLM
judges—Seed2.0Pro (S2P), Seed1.8 (S1.8), Seed1.6 (S1.6)—and two human
groups.}
\label{tab:composition_sft}
\begin{tabular}{lcccccc}
\toprule
\textbf{Setting} & \textbf{IoU} & \textbf{S2P} & \textbf{S1.8} & \textbf{S1.6} & \textbf{User} & \textbf{Expert} \\
\midrule
No comp.   & 0.7484 & 34\% & 34\% & 31\% & 30\%  & 24\%  \\
With comp. & \textbf{0.7529} & \textbf{66\% } & \textbf{66\% } & \textbf{69\%} & \textbf{70\% } & \textbf{76\% } \\
\bottomrule
\end{tabular}
\end{table}

\begin{table*}[h]
\centering
\small
\caption{Comparison between continued Stage~I SFT and Stage~II RL from the same Stage~I checkpoint trained for 10 epochs. Comp-ACC denotes strict composition accuracy. Exp-Cons measures the consistency between the generated explanation and the predicted composition type.}
\label{tab:stage2_necessity}
\resizebox{0.98\textwidth}{!}{
\begin{tabular}{lcccccccc}
\toprule
Method & IoU $\uparrow$ & BDE $\downarrow$ & Comp-ACC $\uparrow$ & Exp-Cons $\uparrow$ & METEOR $\uparrow$ & IoU $>0.5$ & IoU $>0.7$ & IoU $>0.9$ \\
\midrule
Stage~I (10 epochs) & 0.7532 & 0.0509 & 0.7404 & 0.3873 & 0.5129 & 6133 & 4664 & 1268 \\
Stage~I + 5 more epochs SFT & 0.7548 & 0.0506 & 0.7411 & 0.3875 & 0.5126 & 6155 & 4688 & 1290 \\
Stage~II RL (5 epochs) & \textbf{0.7765} & \textbf{0.0447} & \textbf{0.7448} & \textbf{0.3885} & \textbf{0.5194} & \textbf{6263} & \textbf{5017} & \textbf{1507} \\
\bottomrule
\end{tabular}
}
\end{table*}

\subsubsection{Generalization to Real-World Photos (FCDB)}

Since COMEX relies on synthetically outpainted context images, a
natural question is whether the cropping ability learned from it
transfers to real-world photographs.
Table~\ref{tab:fcdb_results} evaluates this on FCDB, a widely used
benchmark of real photos with human-annotated crops.

General-purpose MLLMs again perform poorly, confirming that aesthetic
cropping remains challenging for off-the-shelf vision-language models.
In contrast, our model trained exclusively on COMEX demonstrates
promising zero-shot transferability despite the synthetic-to-real
domain gap. Without any FCDB-specific supervision, Stage~I achieves
an IoU of 0.5946, and Stage~I+II further improves to 0.6251 IoU and
0.0938 BDE, approaching early supervised baselines such as DIC
(0.6300 IoU) that are directly trained on FCDB. This
indicates that the compositional knowledge captured by COMEX
generalizes beyond synthetic images. Notably, GRPO training
(Stage~I+II) yields a +0.0305 IoU gain over Stage~I alone,
suggesting that reinforcement learning on COMEX strengthens not only
in-domain accuracy but also out-of-domain generalization.

We further verify the effectiveness of our two-stage training
pipeline on real-world data by training directly on FCDB from the
base Qwen3-VL-2B, without COMEX pre-training. Stage~I alone already
reaches 0.7007 IoU and 0.0743 BDE, competitive with CACNet (0.7020)
and InstructCrop (0.7100). With the addition of GRPO in Stage~I+II,
performance further improves to \textbf{0.7225} IoU and
\textbf{0.0676} BDE, establishing a new state of the art and
outperforming InstructCrop~(exp) by +0.0065 in IoU. These results
confirm that the benefit of our approach is not limited to the COMEX
setting: the SFT-then-GRPO framework is broadly effective for
aesthetic cropping, and the GRPO stage consistently brings meaningful
gains regardless of the training data source.

\subsection{Ablation Studies}

\subsubsection{Is Composition Information Useful?}



To assess the value of explicit composition supervision, we compare a conventional \textit{crop-and-explanation} setting, which lacks composition classification and explicit composition cues in the explanations, with our proposed \textit{crop-composition-explanation} setting. Crop localization is evaluated using IoU, and explanation quality is assessed by pairwise preference judgments from three MLLM evaluators (Seed2.0Pro, Seed1.8, and Seed1.6). We also conduct a human study on 100 randomly sampled test images with 71 general users and 10 experts. Each score represents the percentage of cases in which evaluators judge one setting to produce a better explanation than the other.

As shown in Table~\ref{tab:composition_sft}, introducing explicit
composition supervision leads to consistent improvements in both crop
localization and explanation quality. IoU improves from 0.7484 to
0.7529, indicating that composition information provides a useful
structural cue for identifying aesthetically favorable crop regions.
More notably, gains in explanation quality are substantial: the
composition-supervised model is preferred by all three MLLM evaluators
with win rates exceeding 66\%, and the advantage is even more
pronounced in human evaluation, where 70\% of general users and 76\%
of experts favor the composition-grounded explanations. This confirms
that composition is not only beneficial for predicting better crops,
but also essential for generating explanations that are more faithful,
interpretable, and grounded in aesthetic principles.

\subsubsection{Is Stage~II Reinforcement Learning Necessary?}

To evaluate whether Stage~II is necessary beyond a strong Stage~I
initialization, we start from the same Stage~I checkpoint trained for
10 epochs and compare two strategies: continuing SFT for 5 additional
epochs or switching to Stage~II reinforcement learning for 5 epochs.

As reported in Table~\ref{tab:stage2_necessity}, simply extending
Stage~I yields only marginal improvements. Continued SFT increases
IoU from 0.7532 to 0.7548, reduces BDE from 0.0509 to 0.0506, and
improves Comp-ACC slightly from 0.7404 to 0.7411, while explanation
metrics remain nearly unchanged. This indicates that Stage~I quickly
saturates once stable structured generation has been learned.

In contrast, Stage~II reinforcement learning brings consistent gains
across all metrics. It improves IoU to 0.7765 and reduces BDE to
0.0447, while also raising Comp-ACC to 0.7448 and METEOR to 0.5194.
The same trend is reflected in the distribution of high-quality crops:
compared with continued SFT, Stage~II produces substantially more
samples with IoU \(> 0.5\), \(> 0.7\), and \(> 0.9\). These results
show that the advantage of Stage~II does not come from longer training
alone, but from task-aligned optimization under explicit reward
signals.

\subsubsection{Is $R_{\mathrm{sem}}$ Necessary?}

\begin{table}[t]
\centering
\small
\setlength{\tabcolsep}{8pt}
\renewcommand{\arraystretch}{1.12}
\caption{Ablation on reward design in Stage~II.}
\label{tab:reward_ablation}
\begin{tabular}{lccc}
\toprule
\textbf{Reward Setting}
  & \textbf{IoU} $\uparrow$
  & \textbf{Comp-ACC} $\uparrow$
  & \textbf{METEOR} $\uparrow$ \\
\midrule
With $R_{\mathrm{sem}}$
  & 0.7765
  & \textbf{0.7448}
  & \textbf{0.5194} \\
Without $R_{\mathrm{sem}}$
  & \textbf{0.7778}
  & 0.7223
  & 0.5060 \\
\bottomrule
\end{tabular}
\end{table}

Table~\ref{tab:reward_ablation} isolates the contribution of the
semantic explanation reward $R_{\mathrm{sem}}$. Removing it and
retaining only the box regression reward $R_{\mathrm{box}}$ yields a
marginal IoU gain of 0.13\%, but composition accuracy drops by 2.25
percentage points and METEOR falls by 1.34 points. The trade-off is
clearly asymmetric: the spatial signal alone pushes the model toward
geometrically accurate crops yet provides no incentive to preserve
compositional reasoning, causing both composition recognition and
explanation generation to degrade. Adding $R_{\mathrm{sem}}$ recovers
these two dimensions at virtually no spatial cost, confirming that
explicit semantic supervision is necessary for the model to produce
explanations that are not only fluent but also faithfully grounded in
composition knowledge.

\subsection{Qualitative Visualization}

\begin{figure}[t]
    \centering
    \includegraphics[width=1\linewidth]{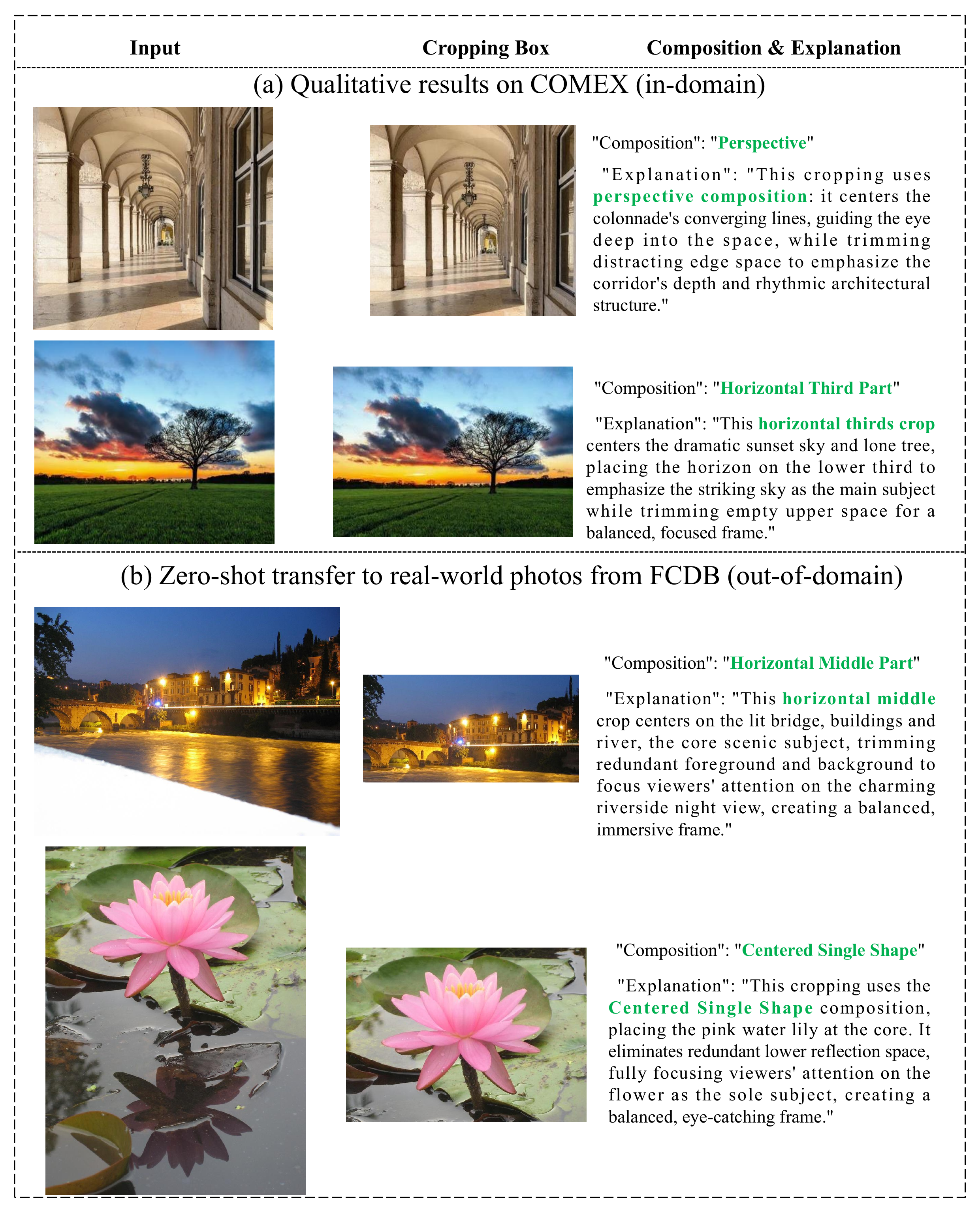}
    \caption{Qualitative cropping results.}
    \label{fig:vis_case}
\end{figure}

Figure~\ref{fig:vis_case} visualizes representative cropping results. On in-domain COMEX examples (Figure~\ref{fig:vis_case}a), the model accurately identifies the crop region and provides composition-grounded explanations across diverse scenes, showing that the two-stage training pipeline effectively equips the MLLM with composition-aware cropping ability. More importantly, when the model trained on COMEX is directly applied to real-world FCDB photographs without any fine-tuning (Figure~\ref{fig:vis_case}b), it still produces visually plausible crops that preserve the salient compositional structure of each scene, such as retaining the central bridge and buildings in the night-view example and tightly centering the lotus. It also generates convincing explanations consistent with the corresponding compositional intent. This suggests that COMEX provides transferable compositional knowledge and helps narrow the gap between synthetic data and real photographs.

\section{Conclusion}

We present a composition-centered formulation of explainable aesthetic image cropping that unifies crop localization, composition prediction, and explanation generation. To support this task, we build COMEX, a large-scale benchmark for composition-grounded cropping and explanation, and propose a two-stage SFT-then-GRPO framework that progressively improves all three outputs. Experiments show strong performance across evaluation metrics, and transfer results on FCDB further demonstrate generalization to real-world photographs.

\noindent\textbf{Limitations.}
COMEX relies on synthetic outpainting and proprietary MLLM-generated annotations, which may introduce domain bias and annotation noise. Future work will incorporate human-annotated real images for more rigorous evaluation.

\bibliographystyle{ACM-Reference-Format}
\bibliography{samples/ref}


\end{document}